\def\paperlanguage{}
\pdfoutput=1 % for arxiv

\documentclass[letterpaper, 10 pt, conference]{ieeeconf}  % Comment this line out if you need a4paper

\usepackage{bm}
\usepackage{cite}
\usepackage{flushend}
\include{preamble}

\newcommand{\ctext}[1]{\raise0.2ex\hbox{\textcircled{\scriptsize{#1}}}}

\IEEEoverridecommandlockouts                              % This command is only needed if
\title{\LARGE \textbf
  {
    \switchlanguage%
    {%
      MEVITA: Open-Source Bipedal Robot Assembled from\\E-Commerce Components via Sheet Metal Welding
    }%
    {%
      MEVITA: ECサイトでの板金溶接により誰でも簡単に構築可能な\\オープンソース二足歩行ロボット
    }%
  }
}

\author{Anonymous authors$^{1}$
  \thanks{$^{1}$ anonymous authors' affiliation.
  }
}

\author{Kento Kawaharazuka$^{*1}$, Shogo Sawaguchi$^{*1}$, Ayumu Iwata$^{1}$,\\Keita Yoneda$^{1}$, Temma Suzuki$^{1}$, and Kei Okada$^{1}$% <-this % stops a space
  \thanks{$^{*}$ K. Kawaharazuka and S. Sawaguchi contributed equally to this work.}
  \thanks{$^{1}$ The authors are with the Department of Mechano-Informatics, Graduate School of Information Science and Technology, The University of Tokyo, 7-3-1 Hongo, Bunkyo-ku, Tokyo, 113-8656, Japan.
    {\texttt\small [kawaharazuka, sawaguchi, a-iwata, k-yoneda, t-suzuki, k-okada]@jsk.imi.i.u-tokyo.ac.jp}
  }%
  }

\begin{document}

\maketitle
\thispagestyle{empty}
\pagestyle{empty}

%%%%%%%%%%%%%%%%%%%%%%%%%%%%%%%%%%%%%%%%%%%%%%%%%%%%%%%%%%%%%%%%%%%%%%%%%%%%%%%%
\begin{abstract}
  \switchlanguage%
  {%
    Various bipedal robots have been developed to date, and in recent years, there has been a growing trend toward releasing these robots as open-source platforms.
    This shift is fostering an environment in which anyone can freely develop bipedal robots and share their knowledge, rather than relying solely on commercial products.
    However, most existing open-source bipedal robots are designed to be fabricated using 3D printers, which limits their scalability in size and often results in fragile structures.
    On the other hand, some metal-based bipedal robots have been developed, but they typically involve a large number of components, making assembly difficult, and in some cases, the parts themselves are not readily available through e-commerce platforms.
    To address these issues, we developed MEVITA, an open-source bipedal robot that can be built entirely from components available via e-commerce.
    Aiming for the minimal viable configuration for a bipedal robot, we utilized sheet metal welding to integrate complex geometries into single parts, thereby significantly reducing the number of components and enabling easy assembly for anyone.
    Through reinforcement learning in simulation and Sim-to-Real transfer, we demonstrated robust walking behaviors across various environments, confirming the effectiveness of our approach.
    All hardware, software, and training environments can be obtained from \href{https://github.com/haraduka/mevita}{\textcolor{magenta}{github.com/haraduka/mevita}}.
    }%
  {%
    これまで様々な二足歩行ロボットが開発されてきており, 近年はこれらをオープンソースとして公開する動きが加速している.
    企業から購入するだけではなく, 誰もが自由に二足歩行ロボットを開発し, その知を共有できるような環境が整いつつある.
    一方で, これまでのオープンソース二足歩行ロボットは, そのほとんどが3Dプリンタによる造形を前提としており, 大きさがスケールしにくく壊れやすいという問題がある.
    また, 一部開発されている金属製の二足歩行ロボットは部品点数が多く, 組み立てが難しかったり, そもそも部品がECサイトで購入できなかったりする.
    そこで本研究では, 全ての部品をe-commerceによって構築可能なオープンソース二足歩行ロボットMEVITAを開発した.
    二足歩行ロボットにおける最小構成を目指しつつ, 板金溶接を用いて複雑な形状を一部品として構築し, 部品点数を大幅に削減することで, 誰でも容易に組み立てられるような設計を実現した.
    シミュレーション上の強化学習とSim-to-Realにより様々な環境における歩行動作を実現し, その有効性を確認した.
    全てのハードウェア・ソフトウェア・学習環境は\url{https://drive.google.com/drive/folders/1PlmsZT_c0rvLUOtIdCrQJ9vXPVhtB7AJ?usp=sharing}に含まれている(採択後githubにおいて公開する).
  }%
\end{abstract}

\section{INTRODUCTION}\label{sec:introduction}
\switchlanguage%
{%
  A wide range of bipedal robots has been developed over the years \cite{hirai1998asimo, kaneko2004hrp2, parmiggiani2012icub, tsagarakis2013coman, englsberger2014toro, tsagarakis2017walkman, seiwald2021lola, liu2022bruce, saloutos2023mithumanoid, unitree2024g1, liao2024berkeley}.
  Traditionally, these robots have relied on harmonic drives as their primary gear mechanism \cite{hirai1998asimo, kaneko2004hrp2, parmiggiani2012icub, tsagarakis2013coman, englsberger2014toro, tsagarakis2017walkman, seiwald2021lola}.
  However, with recent advancements in motor technology, actuators with low gear reduction ratios have become mainstream \cite{liu2022bruce, saloutos2023mithumanoid, unitree2024g1, liao2024berkeley}.
  Such low-gear-ratio robots have made it easier to bridge the gap between simulation and the real world, enabling the application of reinforcement learning to achieve stable walking behaviors \cite{unitree2024g1, radosavovic2024real, tang2024humanmimic}.

  In parallel, a variety of learning methods -- including imitation learning, reinforcement learning, and foundation model-based approaches -- have recently been released as open-source software, creating an environment in which anyone can easily apply these learning techniques.
  This open-source wave is also extending into robotic hardware.
  In recent years, several open-source hardware platforms have been developed for quadrupedal robots \cite{joonyoung2021pawdq, grimminger2020solo, leziart2021solo12, kawaharazuka2024mevius} and bipedal robots \cite{lapeyre2014poppy, allgeuer2015igus, daneshmand2021variable, huang2024stride, xia2024duke, chi2025berkeleylite}.
  While some of these robots provide access to their design files \cite{ficht2017nimbro, shi2025toddlerbot}, they do not fully adhere to the formal definition of open-source \cite{osd2025definition}.
}%
{%
  これまで, 様々な二足歩行ロボットが開発されてきた\cite{hirai1998asimo, kaneko2004hrp2, parmiggiani2012icub, tsagarakis2013coman, englsberger2014toro, tsagarakis2017walkman, seiwald2021lola, liu2022bruce, saloutos2023mithumanoid, unitree2024g1, liao2024berkeley}.
  それらは, これまでハーモニックドライブをギアとして活用していたが\cite{hirai1998asimo, kaneko2004hrp2, parmiggiani2012icub, tsagarakis2013coman, englsberger2014toro, tsagarakis2017walkman, seiwald2021lola}, 現在はモータの発展とともに低減速比なギアを用いたアクチュエータが主流となってきている\cite{liu2022bruce, saloutos2023mithumanoid, unitree2024g1, liao2024berkeley}.
  これら低減速比なロボットはシミュレーションと実機が一致しやすく, 強化学習を用いた歩行動作の実現が可能となってきている\cite{unitree2024g1, radosavovic2024real, tang2024humanmimic}.

  また, 近年摸倣学習や強化学習, 基盤モデルを含む様々な学習手法がオープンソースとして発表され, 誰でも簡単にそれら学習手法を適用できるような環境が整いつつある.
  同時に, その波はロボットのハードウェアにも訪れている.
  近年, 様々なオープンソースハードウェアの四足歩行ロボット\cite{joonyoung2021pawdq, grimminger2020solo, leziart2021solo12, kawaharazuka2024mevius}や二足歩行ロボット\cite{lapeyre2014poppy, allgeuer2015igus, daneshmand2021variable, huang2024stride, xia2024duke, chi2025berkeleylite}が開発されてきている.
  なお, 一部のロボットはハードウェアの設計データが公開されているが\cite{ficht2017nimbro, shi2025toddlerbot}, オープンソースの定義\cite{osd2025definition}からは外れている.
}%

\begin{figure}[t]
  \centering
  \includegraphics[width=0.95\columnwidth]{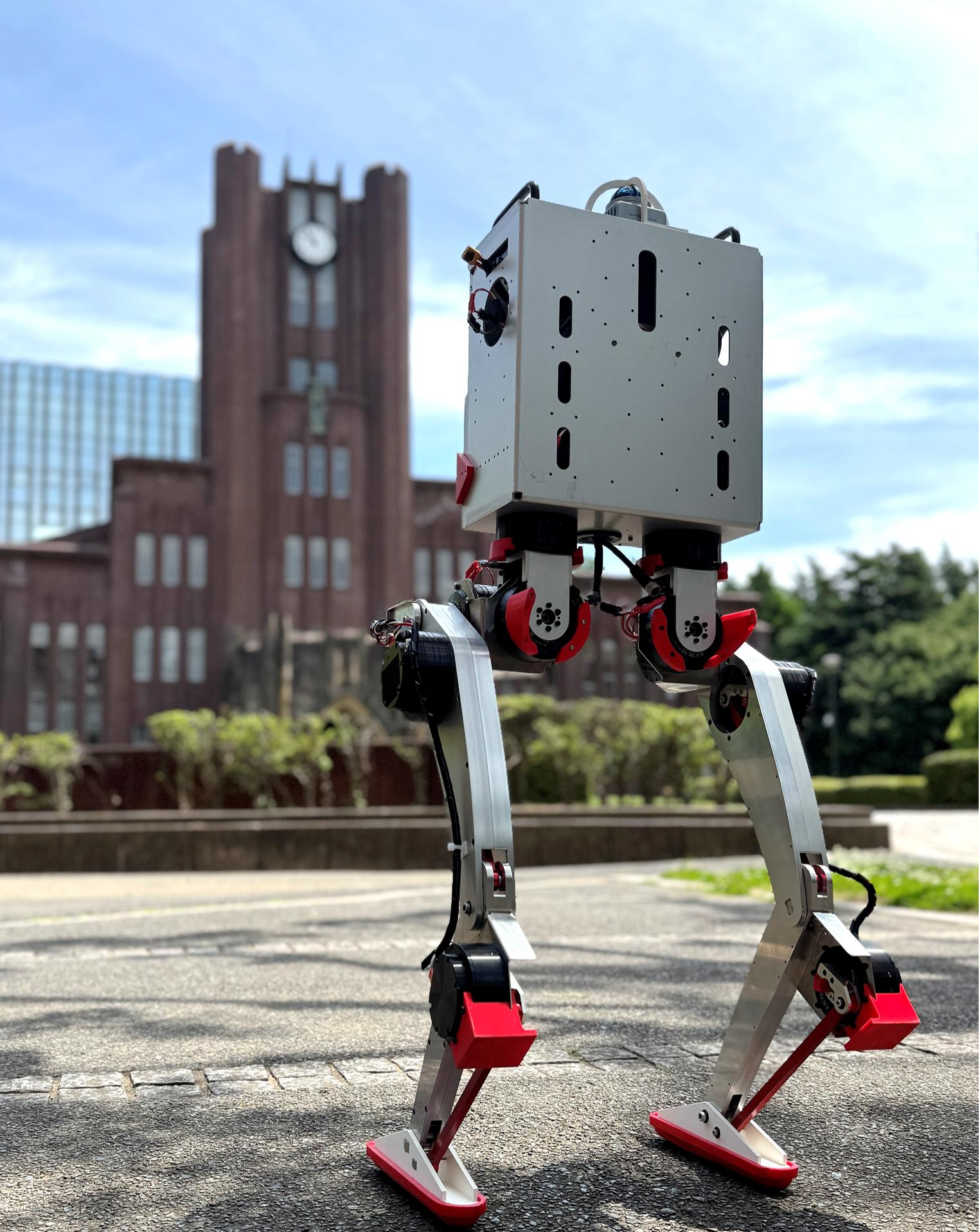}
  % \vspace{-1.0ex}
  \caption{MEVITA -- Open-source bipedal robot easily constructed through e-commerce with sheet metal welding and machining, developed in this study.}
  \label{figure:mevita}
  % \vspace{-3.0ex}
\end{figure}

\begin{figure*}[t]
  \centering
  \includegraphics[width=1.95\columnwidth]{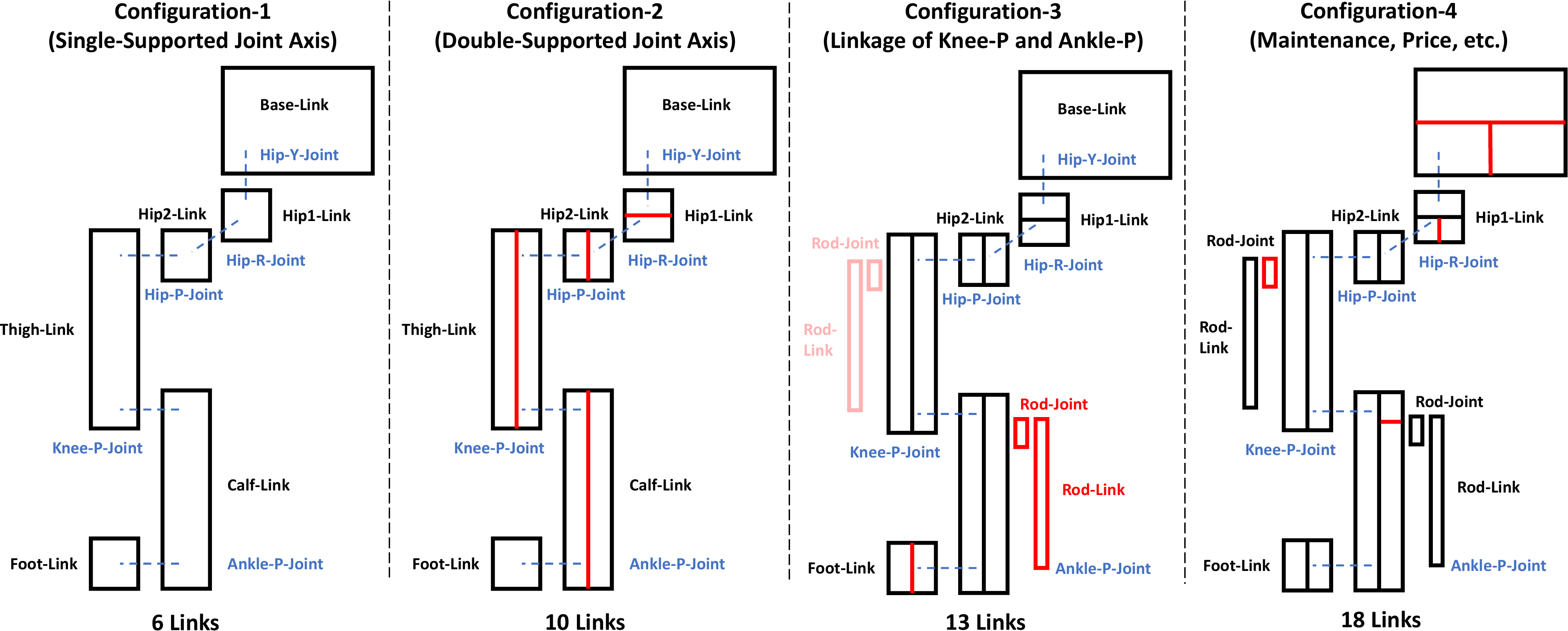}
  % \vspace{-1.0ex}
  \caption{The number of link components in a bipedal robot's various minimal configurations: Configuration-1 represents the case where all joints are single-supported; Configuration-2 assumes all joints except the Hip-Y joint are double-supported; Configuration-3 introduces parallel linkages at the Knee-P and Ankle-P joints; and Configuration-4 corresponds to the design adopted in MEVITA, taking into account factors such as maintenance and cost.}
  \label{figure:minimum-design}
  % \vspace{-2.0ex}
\end{figure*}

\switchlanguage%
{%
  In this study, we focus specifically on open-source bipedal robots.
  Most existing open-source bipedal robots \cite{lapeyre2014poppy, allgeuer2015igus, daneshmand2021variable, chi2025berkeleylite} are designed with 3D printing in mind.
  While 3D printing offers advantages in terms of ease of fabrication, cost, and geometric flexibility, it poses limitations in terms of structural strength, making it difficult to scale up or enable dynamic, high-intensity motions.
  In contrast, a few metal-based open-source bipedal robots have also been developed \cite{huang2024stride, xia2024duke}.
  However, due to the reliance on traditional metal machining, these robots often consist of many components, making them difficult to assemble.
  Additionally, the required parts are not always available through e-commerce platforms.

  To address these issues, we developed MEVITA, an open-source bipedal robot whose entire frame is composed of metal and whose mechanical and electronic components can be fully sourced from e-commerce platforms.
  In particular, by utilizing MISUMI and its online manufacturing service meviy \cite{misumi2024meviy}, we enabled the procurement of not only bearings and shafts but also all structural links and joint components directly from e-commerce platforms.
  Furthermore, we propose a design principle based on minimizing the number of components, from the minimal viable configuration to a fully functional bipedal robot.
  Although designing freeform geometries is more challenging with metal compared to 3D printing, we addressed this by employing sheet metal welding to integrate complex geometries into single components.
  As a result, MEVITA achieves a significant reduction in the number of parts compared to previous robots, allowing for an easily assemblable design.
  Excluding mirrored components, MEVITA's skeletal frame consists of only 18 unique metal parts.
  Despite its simple CAN-based circuit architecture, MEVITA is capable of robust walking across diverse environments, enabled by reinforcement learning in simulation and Sim-to-Real transfer.
}%
{%
  本研究では, その中でも特にオープンソースの二足歩行ロボットに着目した.
  これまでのオープンソースの二足歩行ロボット\cite{lapeyre2014poppy, allgeuer2015igus, daneshmand2021variable, chi2025berkeleylite}は, そのほとんどが3Dプリンタによる造形を前提としている.
  もちろん3Dプリンタは作りやすさや値段, 形状の自由度の高さにおいて非常に優れているが, 強度の問題からスケールを大きくしたり, 激しい動きをさせたりすることは難しい.
  これに対して, 金属で構成されたオープンソースの二足歩行ロボットも一部開発されている\cite{huang2024stride, xia2024duke}.
  しかし, 金属切削を考えた結果として部品点数が多く, 組み立てが難しかったり, そもそも部品がECサイトで購入できなかったりする.

  そこで本研究では, 全体の骨格が金属であり, 回路や機械部品の全てをe-commerceサイトから構築できるオープンソース二足歩行ロボットMEVITAを開発した.
  特に, e-commerceサイトであるMISUMIとその加工サービスであるmeviy \cite{misumi2024meviy}を利用することで, ベアリングや軸だけでなく, 全てのリンク・関節部品をe-commerceサイトから構築できる.
  また, 二足歩行ロボットの最小構成から合理的な構成まで, 部品点数最小化に基づく設計原理を考えた.
  3Dプリントに比べて自由形状の難しい金属部品における部品点数の上昇については, 板金溶接を用いて複雑な形状を一部品として構築することで対処した.
  これにより, 部品点数をこれまでのロボットよりも大幅に削減し, 容易に組み立てられるような設計を実現した.
  ミラー部品を除くとMEVITAの骨格を構成する部品は全部で18種類しかない.
  MEVITAはCANによるシンプルな回路構成を持ちつつ, シミュレーション上における強化学習とSim-to-Realにより, 多様な環境における歩行動作を実現可能である.

  % 本研究の構成について述べる.
  % \secref{sec:mevita}では, MEVITAの部品点数最小化に基づく設計原理, MEVITAの身体設計の概要, 他ロボットとの比較, MEVITAの回路構成, 強化学習を用いた制御について述べる.
  % \secref{sec:experiment}では, MEVITAにおける多様な環境での歩行実験を行い, その有効性を確認する.
  % \secref{sec:discussion}でMEVITAの設計上の課題について議論し, \secref{sec:conclusion}で結論を述べる
}%

\begin{figure*}[t]
  \centering
  \includegraphics[width=1.95\columnwidth]{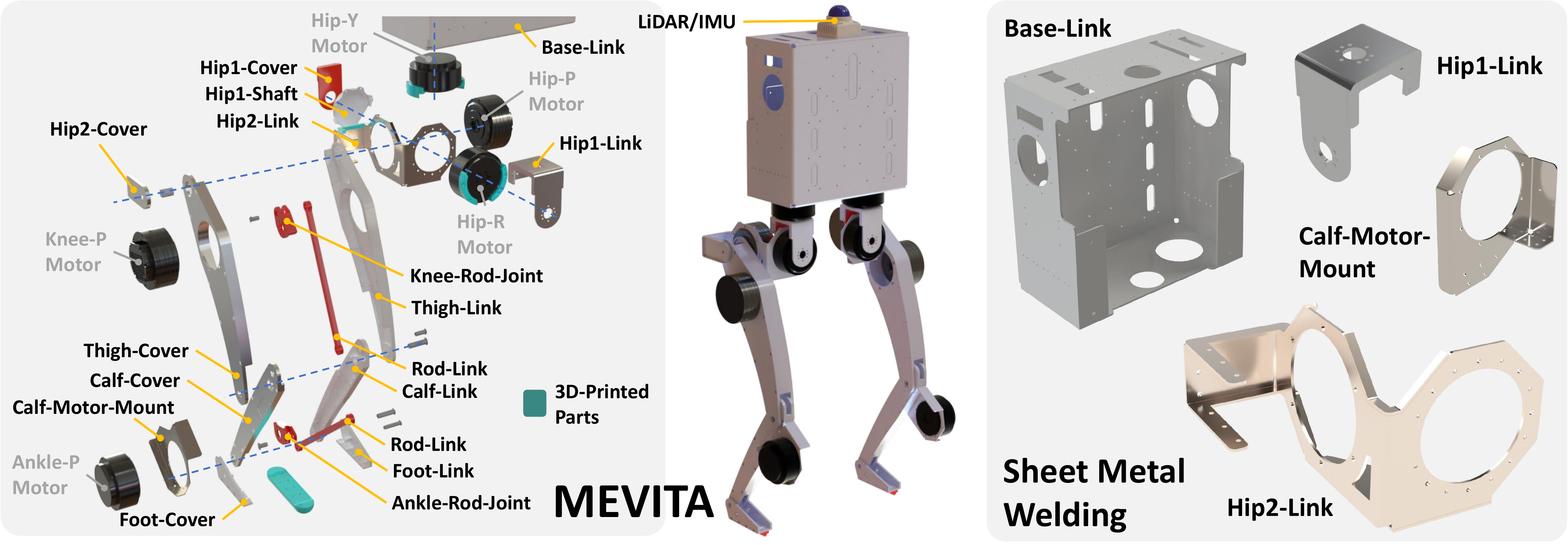}
  % \vspace{-1.0ex}
  \caption{Design overview of MEVITA: excluding mirrored parts, the robot consists of 18 unique metal components, four of which are fabricated using sheet metal welding to achieve complex and large geometries as single integrated parts.}
  \label{figure:mevita-design}
  % \vspace{-1.0ex}
\end{figure*}

\begin{figure}[t]
  \centering
  \includegraphics[width=0.95\columnwidth]{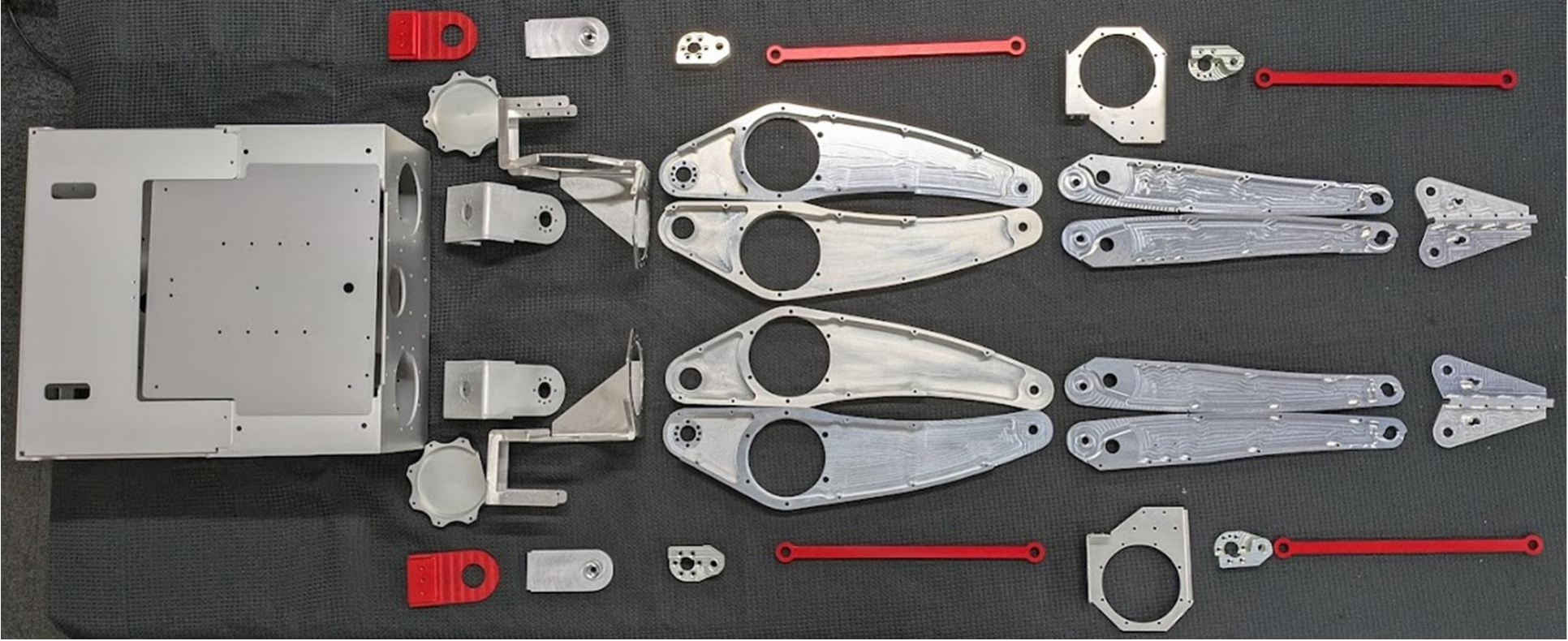}
  % \vspace{-1.0ex}
  \caption{All metal components used in MEVITA.}
  \label{figure:mevita-parts}
  % \vspace{-3.0ex}
\end{figure}

\begin{figure}[t]
  \centering
  \includegraphics[width=0.9\columnwidth]{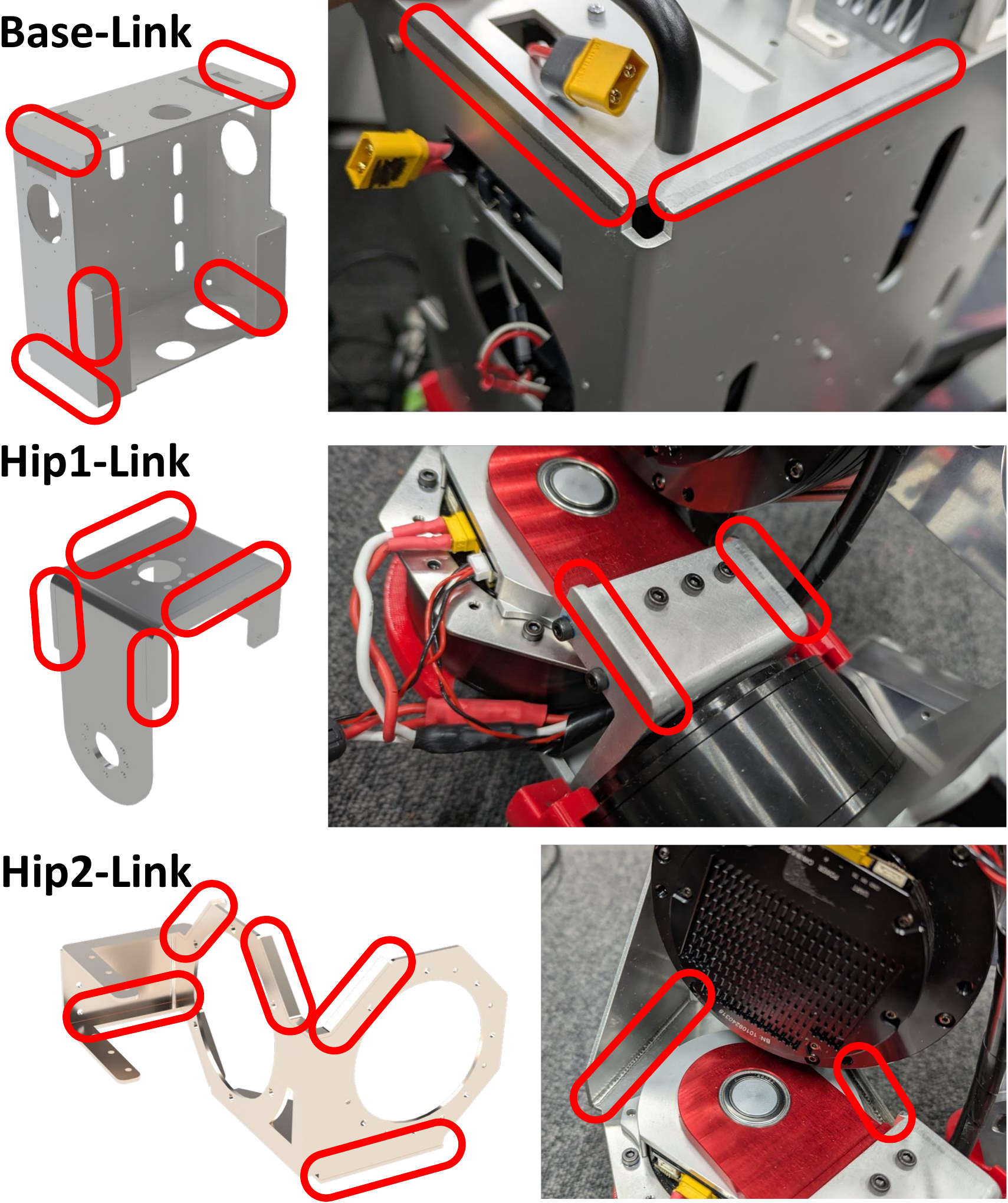}
  % \vspace{-1.0ex}
  \caption{Details of sheet metal welding for the Base-Link, Hip1-Link, and Hip2-Link. The welded sections are highlighted in red.}
  \label{figure:mevita-welding}
  % \vspace{-2.0ex}
\end{figure}

\section{Design and Configuration of MEVITA} \label{sec:mevita}

\subsection{Minimum Design of Bipedal Robots} \label{subsec:minimum-design}
\switchlanguage%
{%
  First, we consider the minimal configuration of a bipedal robot.
  In this study, we focus on a 5-DoF leg configuration that excludes the ankle roll joint, following the design of robots such as Duke Humanoid \cite{xia2024duke}, Cassie \cite{reher2019cassie}, and MIT Humanoid \cite{chignoli2021mithumanoid}.
  Each leg, therefore, consists of a 3-DoF hip joint, a 1-DoF knee joint, and a 1-DoF ankle joint.
  The order of the three DoFs in the hip joint can vary depending on the robot, with examples such as Y-R-P \cite{hirai1998asimo, kaneko2004hrp2, xia2024duke, chignoli2021mithumanoid}, R-Y-P \cite{reher2019cassie}, and P-R-Y \cite{unitree2024g1}, where R, P, and Y denote roll, pitch, and yaw joints, respectively.
  In this work, we adopt the most widely used Y-R-P order.
  Note that the number of components required for the minimal configuration is unaffected by the chosen joint order.

  The left side of \figref{figure:minimum-design} shows Configuration-1, which represents the minimal structure of a bipedal robot.
  A robot with 5-DoF per leg and a Y-R-P hip structure has the following joints: Hip-Y, Hip-R, Hip-P, Knee-P, and Ankle-P
  To connect these joints, six links are needed: Base, Hip1, Hip2, Thigh, Calf, and Foot.
  Thus, considering just one leg and the torso, six links form the minimal configuration.

  Configuration-2 builds upon Configuration-1 by converting the single-supported (cantilevered) joints into more practical double-supported structures.
  Since the Hip-Y joint generally experiences less torque, it is common to apply double support only to the Hip-R, Hip-P, Knee-P, and Ankle-P joints.
  As a result, the Hip1, Hip2, Thigh, and Calf links become two-part assemblies, bringing the total to ten distinct link components in the minimal double-supported configuration.

  In recent bipedal robot designs, to increase leg agility, Configuration-3 introduces parallel link mechanisms at the Knee-P and Ankle-P joints.
  This allows the motors to be positioned more proximally, thereby reducing the link's moment of inertia.
  To implement this, two additional components are required: a Rod-Joint attached to the motor and a parallel Rod-Link forming the linkage (ideally shared for both Knee-P and Ankle-P).
  Furthermore, to implement a double-supported Rod-Link, the Foot-Link must be split into two parts (whereas the split in the Calf-Link from earlier can be reused for the Knee-P joint).
  Although the Rod-Joint should also be split for double support, since it is a small component mounted at the motor's tip, it can be designed as a single piece while maintaining double support.
  Therefore, Configuration-3, which includes both double-supported structures and parallel links, results in a minimal configuration consisting of 13 distinct link components.

  Finally, considering factors like maintenance and material cost, MEVITA adopts Configuration-4, consisting of 18 components. While 13 components represent the theoretical minimum in Configuration-3, realizing this in practice is difficult.
  This is because each individual part becomes very large, resulting in higher costs and reduced maintainability.
  Therefore, in Configuration-4, the Base-Link is split into three parts -- a main body, a cover, and a battery access hatch -- to improve maintainability and ease of battery removal.
  Similarly, the Hip1-Link and Calf-Link are each split into three parts to reduce material costs by keeping individual components smaller.
  To account for the different torque requirements at the Knee-P and Ankle-P joints, we use separate motors for each and thus design the Rod-Joint as two distinct components.
  Consequently, MEVITA's Configuration-4 comprises 18 unique link components.
  On the other hand, even with just these 18 parts, it is quite challenging to fabricate all of them solely through simple machining.
  Large structural components would require extensive material removal if machined from solid blocks, leading to excessive cost and production time.
  In practice, such components are usually subdivided further to facilitate efficient manufacturing.
  In this study, we instead explored the use of sheet metal welding to construct complex shapes as single integrated parts.
}%
{%
  まず, 二足歩行ロボットの最小構成について考える.
  本研究では, Duke Humanoid \cite{xia2024duke}やCassie \cite{reher2019cassie}, MIT Humanoid \cite{chignoli2021mithumanoid}と同様に, 足首のロール関節を除く各脚5自由度の二足歩行ロボットを考える.
  つまり, ロボットは3自由度の腰関節, 1自由度の膝関節, 1自由度の足首関節を持つ.
  これらのロボットにおいて, 腰関節の順番は様々考えられ, Y-R-P \cite{hirai1998asimo, kaneko2004hrp2, xia2024duke, chignoli2021mithumanoid}, R-Y-P \cite{reher2019cassie}, P-R-Y \cite{unitree2024g1}などの順番が用いられている(RはRoll, PはPitch, YはYaw関節を表す).
  ここでは, 最も広く採用されているY-R-Pの順番で考える.
  なお, どれを採用しても最小構成については部品数は変わらない.

  \figref{figure:minimum-design}の左図に, 二足歩行ロボットの最小構成であるConfiguration-1を示す.
  各脚が5自由度であり腰関節がY-R-Pの構造である二足歩行ロボットは, Hip-Y, Hip-R, Hip-P, Knee-P, Ankle-Pの5つの関節軸を持つ.
  これを繋ぐリンクとして, Base, Hip1, Hip2, Thigh, Calf, Footの6つのリンクが必要である.
  つまり, 胴体と片脚だけを考えると, 6つのリンクが最小構成となる.

  次に, 関節軸に対してリンクが片持ちに取り付けられたConfiguration-1を, より実用的な両持ち構造にしたものがConfiguration-2である.
  一般的にHip-Y関節には大きなモーメントがかからないため, Hip-R, Hip-P, Knee-P, Ankle-Pの4つの関節軸を両持ち構造にする場合が多い.
  よって, Hip1, Hip2, Thigh, Calfの4つのリンクが両持ち構造により2部品となり, 10個のリンクが最小構成となる.

  次に, 近年の二足歩行ロボットは脚をより素早く動かすために, Configuration-3のように平行リンク構造をKnee-PとAnkle-Pに適用する場合が多い.
  これにより, モータをより近位に配置でき, リンクの慣性モーメントを小さくすることができる.
  そのため, モータにとりつけるRod-Jointと, 平行リンクをなすRod-Linkがそれぞれ1部品ずつ必要となる(理想的にはKnee-PにもAnkle-Pにも同様のものを利用できる).
  また, このRod-Linkを両持ち構造にするために, Foot-Linkも2部品に分割する必要がある(Knee-PについてはCalf-Linkの分割をそのまま活用できる).
  なお, 本来であればRod-Jointも両持ち構造にするため分割する必要があるが, この部品はモータ先端に取り付く小さな部品であり, うまく構造を作れば1部品でも両持ち構造にすることができる.
  よって, 両持ち構造かつ平行リンクを持つConfiguration-3は, 13個のリンクが最小構成となる.

  最後に, メンテナンスや材料費などを考慮して, MEVITAはConfiguration-4に示すように18部品で構成されている.
  最小構成はConfiguration-3の13部品であるが, 実際にはそれを実現することは容易ではない.
  一つ一つの部品が非常に大きくコストがかさんだり, メンテナンス性が下がったりするためである.
  まずBase-Linkはメンテナンス性向上とバッテリーの取り出しやすさを考慮して, 本体とカバー, そしてバッテリー取り出し口の3部品に分けている.
  また, Hip1-LinkとCalf-Linkは, 材料費を考慮して各部品を小さくするために, それぞれ2部品から3部品に分割している.
  Rod-Jointについても, Knee-PとAnkle-Pに必要なトルクを考慮して別々のモータを利用するため, Rod-Jointを両者で別部品とした.
  よって, MEVITAの構成であるConfiguration-4は, 18部品から構成されている.
  一方で, この18部品でさえ, 単純な切削部品のみで全て実現することはかなり難しい.
  大きな構造を切削により実現する場合, 削り出しの量が増え, 膨大なお金と時間がかかるため, これをさらに細かく分解することが一般的である.
  そこで本研究では, 板金溶接を用いて, 複雑な形状を一部品として構築することを考えた.
}%

\begin{table*}[t]
  \centering
  \caption{Comparison between existing bipedal robots and MEVITA}
  \begin{tabular}{l|ccccccc}
    Name & Weight   & Leg Length\textsuperscript{*1} & Leg DoFs & Materials\textsuperscript{*2} & Open/Closed (CAD) & Number of Metal Parts\textsuperscript{*3}\\ \hline % & Maximum Torque \\ \hline
    Cassie \cite{reher2019cassie}                     & 33.32 kg                  & 0.5 m  &  5 & Metal   & Closed        & -  \\ % & 195 Nm   \\
    MIT Humanoid \cite{saloutos2023mithumanoid}       & 21 kg\textsuperscript{*4} & 0.28 m &  5 & Metal   & Closed        & -  \\ % & 136.0 Nm \\
    Berkeley Humanoid \cite{liao2024berkeley}         & 16 kg                     & 0.2 m  &  6 & Metal   & Closed        & -  \\ % & 81.1 Nm  \\
    Bolt \cite{daneshmand2021variable}                & 1.34 kg                   & 0.2 m  &  3 & Plastic & \textbf{Open} & -  \\ % & 2.5 Nm   \\
    Berkeley Humanoid Lite \cite{chi2025berkeleylite} & 16 kg\textsuperscript{*4} & 0.16 m &  6 & Plastic & \textbf{Open} & -  \\ % & 3.03 Nm  \\
    Duke Humanoid \cite{xia2024duke}                  & 30 kg                     & 0.25 m &  5 & Metal   & \textbf{Open} & 24 \\ % & 264 Nm   \\
    MEVITA (This Study)                               & 19.8 kg                   & 0.32 m &  5 & Metal   & \textbf{Open} & 18 \\ % & 48 Nm    \\
  \end{tabular}
  \label{table:comparison}
  \begin{flushleft}
  \footnotesize
    \textsuperscript{*1} The average of the link lengths of the thigh and calf links. \\
    \textsuperscript{*2} The materials of the functional components forming the robot's skeletal structure.\\
    \textsuperscript{*3} The number of the metal components forming the skeletal structure, excluding the mirror parts.\\
    \textsuperscript{*4} Including the weight of the arms.
  \end{flushleft}
  % \vspace{-3.0ex}
\end{table*}

\subsection{Design Overview of MEVITA} \label{subsec:mevita-design}
\switchlanguage%
{%
  An overview of the MEVITA design is shown in \figref{figure:mevita-design}, and all the metal components used are shown in \figref{figure:mevita-parts}.
  MEVITA features five degrees of freedom per leg.
  The Hip-P and Knee-P joints use T-MOTOR AK10-9 actuators, while the Hip-Y, Hip-R, and Ankle-P joints use T-MOTOR AK70-10.
  All structural components of MEVITA are made from either aluminum or stainless steel.
  Machined metal parts use A7075, standard sheet metal parts are made of A5052, and certain high-strength sheet metal components are fabricated from SUS304.
  Additional parts such as the foot sole and joint limiters are 3D-printed using TPU.
  Each link is divided into a *-Link and *-Cover to achieve a double-supported structure.
  Additionally, for cost reduction in Configuration-4, some new components -- such as the Hip1-Shaft and Calf-Motor-Mount -- have been introduced.
  As for sensors, a Livox Mid-360 is mounted on the top of the Base-Link.
  This sensor not only enables environmental perception and SLAM but also functions as an IMU for control purposes.

  A key feature of MEVITA is that four of its link components -- Base-Link, Hip1-Link, Hip2-Link, and Calf-Motor-Mount -- are constructed via sheet metal welding.
  The Base-Link and Hip1-Link are made from A5052, while the Hip2-Link and Calf-Motor-Mount are made from SUS304.
  As shown in \figref{figure:mevita-welding}, welding multiple sheet metal plates together allows the construction of highly complex shapes as single parts.
  In particular, the Base-Link is a single component that forms the entire torso, measuring 300$\times$320$\times$150 [mm].
  The Hip2-Link is also a single part, formed by welding together four sheets to achieve a complex yet strong structure.
  These components are designed to be compatible with meviy, the manufacturing service provided by MISUMI \cite{misumi2024meviy}, and can be automatically quoted and ordered through e-commerce using only STEP files.
}%
{%
  開発したMEVITAの設計概要を\figref{figure:mevita-design}に, 用いられている全ての金属部品を\figref{figure:mevita-parts}に示す.
  MEVITAは各脚に5自由度を持ち, Hip-PとKnee-PはT-MOTOR AK10-9を, Hip-Y, Hip-R, Ankle-PはT-MOTOR AK70-10を用いている.
  MEVITAの構造をなす部品は全てアルミニウムまたはステンレスの金属により構成されており, 金属切削部品はA7075, 板金部品は基本的にA5052, 一部の板金部品は強度の問題からSUS304を用いている.
  その他の足平や関節リミット部品についてはTPUを用いた3Dプリント部品である.
  各リンクは両持ち構造にするために, *-Linkと*-Coverに分割されており, 一部Configuration-4に向けた材料費削減のために, Hip1-ShaftやCalf-Motor-Mountが追加されている.
  センサとしては, Base-Linkの上面にLivox Mid-360が搭載されており, これにより周囲の環境を認識してSLAMを行うだけでなく, 制御用のIMUとしても活用することができる.

  ここで重要となるのは, Base-Link, Hip1-Link, Hip2-Link, Calf-Motor-Mountの4リンクが板金溶接により構成されていることである.
  Base-LinkとHip1-LinkはA5052, Hip2-LinkとCalf-Motor-MountはSUS304により構成されている.
  \figref{figure:mevita-welding}に示すように, 複数の板金を溶接することで, 非常に複雑な形状を1部品として構築することができる.
  特に, Base-Linkは, たった一部品で300$\times$320$\times$150 [mm]の胴体が構成されている.
  また, Hip2-Linkは, 4枚の板金を溶接することで, 強度を保ちつつ複雑な構造を一部品として構築することに成功している.
  これらの部品はMISUMIの加工サービスであるmeviy \cite{misumi2024meviy}を用いて, STEPファイルのみから自動でe-commerceを通して見積もり・発注することができる形状となっている.
}%

\subsection{Comparison with Existing Bipedal Robots} \label{subsec:comparison}
\switchlanguage%
{%
  A comparison between MEVITA and existing bipedal robots is shown in \tabref{table:comparison}.
  The table compares seven robots, including MEVITA.
  Among them, Cassie \cite{reher2019cassie}, MIT Humanoid \cite{saloutos2023mithumanoid}, and Berkeley Humanoid \cite{liao2024berkeley} are not open-source, while Bolt \cite{daneshmand2021variable}, Berkeley Humanoid Lite \cite{chi2025berkeleylite}, Duke Humanoid \cite{xia2024duke}, and MEVITA are open-source.
  Although the robots differ in weight, leg length, and number of leg DoFs, MEVITA has a leg length and weight relatively close to those of MIT Humanoid and Duke Humanoid.
  Among the open-source robots, Bolt and Berkeley Humanoid Lite are designed with 3D printing in mind.
  MEVITA, like the Duke Humanoid, is based on metal machining, but the number of metal parts in MEVITA has been reduced from 24 (in the Duke Humanoid) to 18.
  Considering the 13-part minimum configuration (Configuration-3) shown in \figref{figure:minimum-design}, the Duke Humanoid adds 11 extra parts, while MEVITA adds only 5.
  This demonstrates that by incorporating sheet metal welding, MEVITA successfully limits the increase in part count to less than half compared to conventional metal-machined designs.
}%
{%
  本研究で開発したMEVITAとこれまでの二足歩行ロボットとの比較を\tabref{table:comparison}に示す.
  ここでは, MEVITAを含めた7つのロボットを比較している.
  Cassie \cite{reher2019cassie}, MIT Humanoid \cite{saloutos2023mithumanoid}, Berkeley Humanoid \cite{liao2024berkeley}はオープンソースではなく, Bolt \cite{daneshmand2021variable}, Berkeley Humanoid Lite \cite{chi2025berkeleylite}, Duke Humanoid \cite{xia2024duke}, MEVITAはオープンソースである.
  どのロボットも重さや脚の長さ, 脚の自由度数は異なるが, MEVITAはMIT HumanoidやDuke Humanoidと比較的近い脚の長さや重さを持っている.
  オープンソースであるロボットのうち, BoltとBerkeley Humanoid Liteは, 3Dプリンタによる造形を前提とている.
  MEVITAは, Duke Humanoidと同じく金属切削を前提としているが, MEVITAはDuke Humanoidに比べて金属部品の点数が24から18に減っている.
  \figref{figure:minimum-design}における最小構成Configuration-3の13部品からの増減を考えると, Duke Humanoidは11部品が追加, MEVITAは5部品が追加されており, 板金溶接の併用により部品点数の増加を半分以下に抑えることに成功している.
}%

\begin{figure}[t]
  \centering
  \includegraphics[width=0.95\columnwidth]{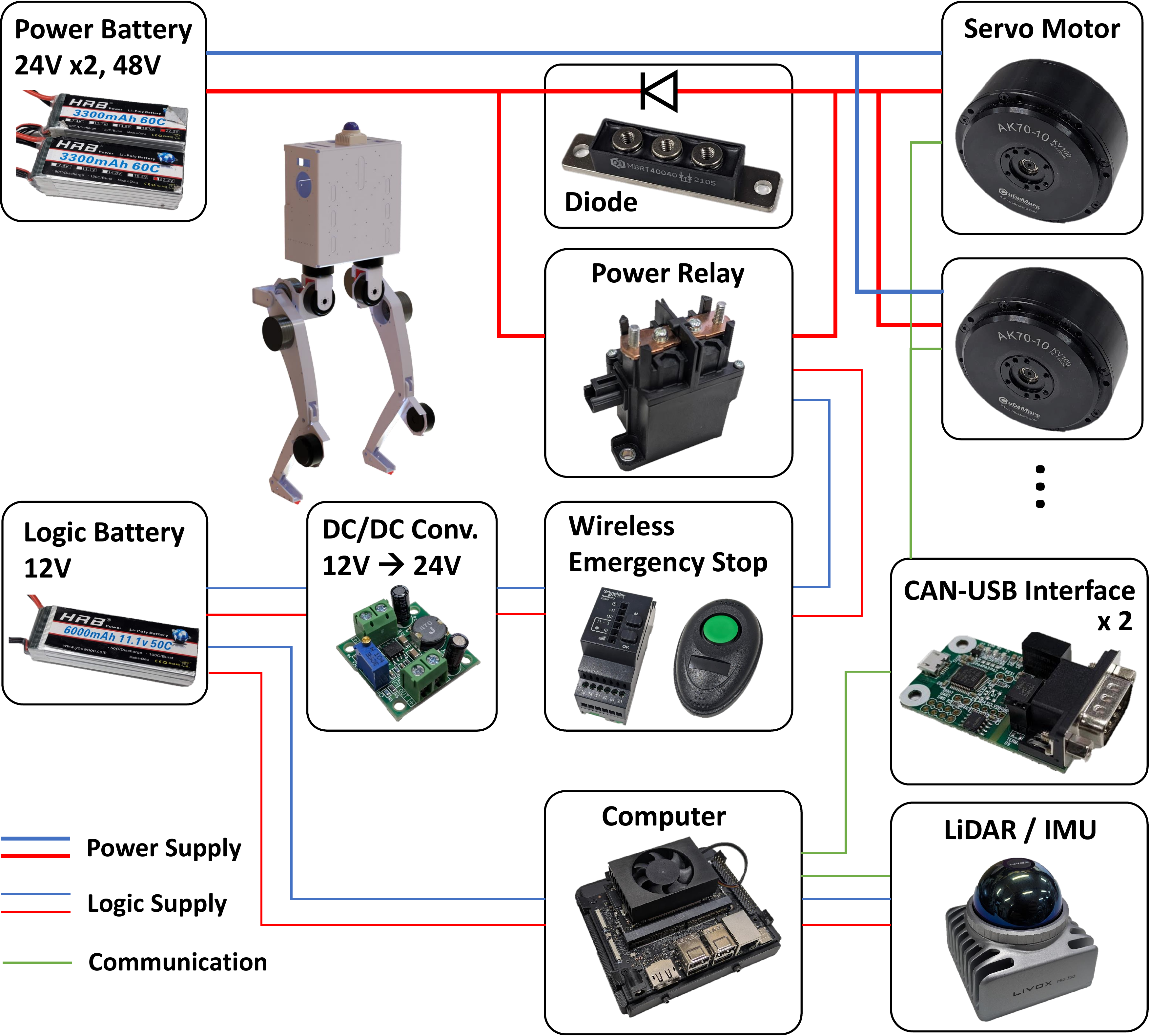}
  % \vspace{-1.0ex}
  \caption{Circuit configuration of MEVITA: servo motors are connected to the PC via two CAN-USB interfaces, and the system is equipped with a wireless emergency stop, power relay, diode, and LiDAR/IMU.}
  \label{figure:mevita-circuit}
  % \vspace{-2.0ex}
\end{figure}

\begin{figure}[t]
  \centering
  \includegraphics[width=0.95\columnwidth]{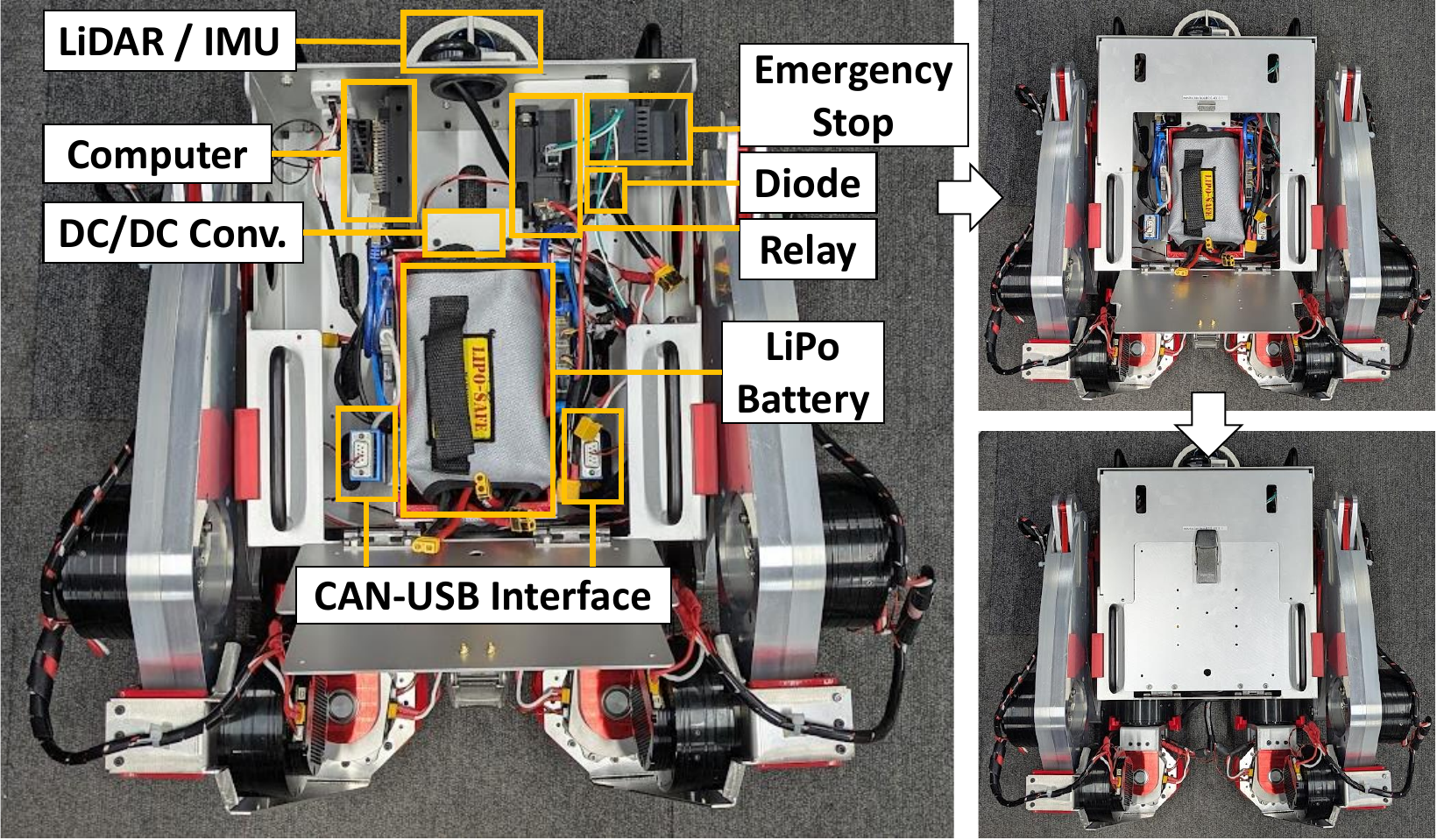}
  % \vspace{-1.0ex}
  \caption{The actual circuit layout inside MEVITA's Base-Link.}
  \label{figure:actual-circuit}
  % \vspace{-1.0ex}
\end{figure}

\subsection{Circuit Configuration of MEVITA} \label{subsec:mevita-circuit}
\switchlanguage%
{%
  The circuit configuration of MEVITA is shown in \figref{figure:mevita-circuit}.
  MEVITA uses the NVIDIA Jetson Orin Nano Developer Kit as its onboard computer.
  Connected to it are two CAN-USB interfaces and a Livox Mid-360, a LiDAR sensor equipped with an IMU.
  Motor power is managed via a wireless emergency stop switch and a power relay.
  MEVITA is powered by LiPo batteries: two 24 V batteries are connected in series to provide 48 V for the motors, while a separate 12 V battery is used for logic power supply.
  The actual internal layout of MEVITA's electronics inside the Base-Link is shown in \figref{figure:actual-circuit}.
  The LiPo battery is centrally located, with other devices arranged around it with sufficient clearance, making future expansion relatively easy.
}%
{%
  MEVITAの回路構成を\figref{figure:mevita-circuit}に示す.
  MEVITAはコンピュータとしてNVIDIA Jetson Orin Nano Developer Kitを用いている.
  これに, 2つのCAN-USBインターフェイス, IMUつきのLidarであるLivox Mid-360を接続している.
  無線緊急停止スイッチとパワーリレーによってモータの電源を管理している.
  LiPoバッテリーを用いており, パワーは24 Vのバッテリーを直列に接続して48 Vとして, ロジックは12 Vのバッテリーを用いている.
  実際のMEVITAのBase-Link内部の回路配置を\figref{figure:actual-circuit}に示す.
  真ん中に配置されたLiPoバッテリーを中心として, 各デバイスが余裕を持って配置されており, 比較的拡張は容易である.
}%

\begin{figure}[t]
  \centering
  \includegraphics[width=0.95\columnwidth]{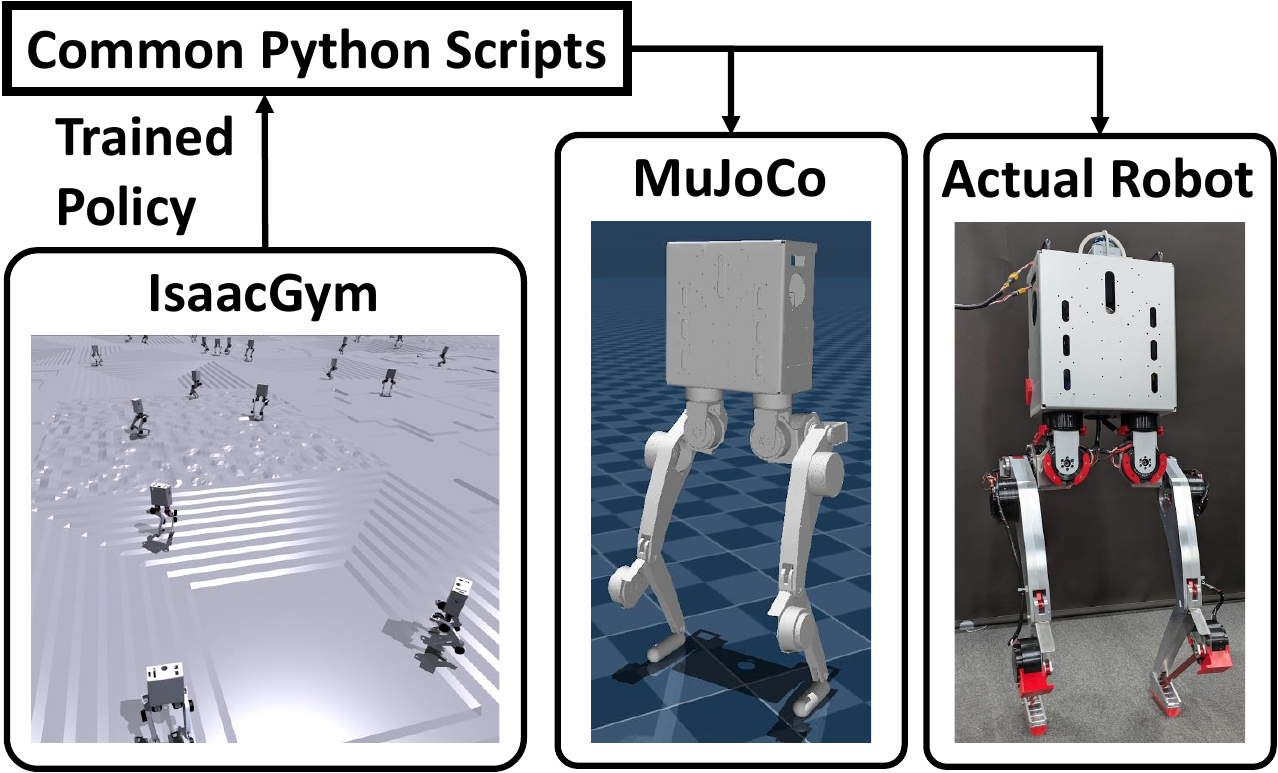}
  % \vspace{-1.0ex}
  \caption{Control system of MEVITA: policies trained in IsaacGym are verified in MuJoCo and deployed to the actual hardware using common Python scripts.}
  \label{figure:mevita-control}
  % \vspace{-2.0ex}
\end{figure}

\subsection{Control Architecture of MEVITA} \label{subsec:mevita-control}
\switchlanguage%
{%
  The control architecture of MEVITA is illustrated in \figref{figure:mevita-control}.
  MEVITA's control is primarily based on reinforcement learning using IsaacGym / LeggedGym \cite{rudin2022leggedgym}.
  The learned policy is first validated through Sim-to-Sim transfer using MuJoCo \cite{todorov2012mujoco}, and then finally tested on the real hardware.
  The control input consists of joint angle commands for PID control.
  The state vector includes the angular velocity and gravity direction vector of the Base-Link, the desired velocity command for the Base-Link, joint positions, and joint velocities.
  The reward function is adapted from those commonly used for quadrupedal locomotion tasks and applied to the bipedal context.
  To enable Sim-to-Real transfer, a variety of domain randomization techniques are employed.
  These include noise added to the mass, moment of inertia, and center of mass of each link, as well as to friction coefficients and control command latency.
  For further details, please refer to the learning environment software available at \href{https://github.com/haraduka/mevita}{\textcolor{magenta}{github.com/haraduka/mevita}}.
}%
{%
  MEVITAの制御アーキテクチャを\figref{figure:mevita-control}に示す.
  MEVITAの制御は基本的にIsaacGym / LeggedGym \cite{rudin2022leggedgym}を用いた強化学習により行い, それをMuJoCo \cite{todorov2012mujoco}を用いたSim-to-Simにより検証, 最後に実機により動作を確認する流れとなっている.
  制御入力としてPID制御の関節角度指令値を用い, 状態としてはBase-Linkの回転速度と重力方向ベクトル, Base-Linkの速度指令値, 関節角度, 関節角速度を用いたシンプルな構成である.
  報酬関数は一般的な四脚ロボットの歩行に用いられている報酬関数を二足歩行ロボットに適用したものである.
  また, Sim-to-Realのために, 各リンクの質量, 慣性モーメント, 重心位置, 摩擦係数や, 制御指令値の遅れなど, 様々なランダムノイズを加えている.
  詳細は\url{https://drive.google.com/drive/folders/1PlmsZT_c0rvLUOtIdCrQJ9vXPVhtB7AJ?usp=sharing}に含まれている学習環境のソフトウェア(採択後githubにおいて公開する)を参照されたい.
}%

\begin{figure}[t]
  \centering
  \includegraphics[width=0.95\columnwidth]{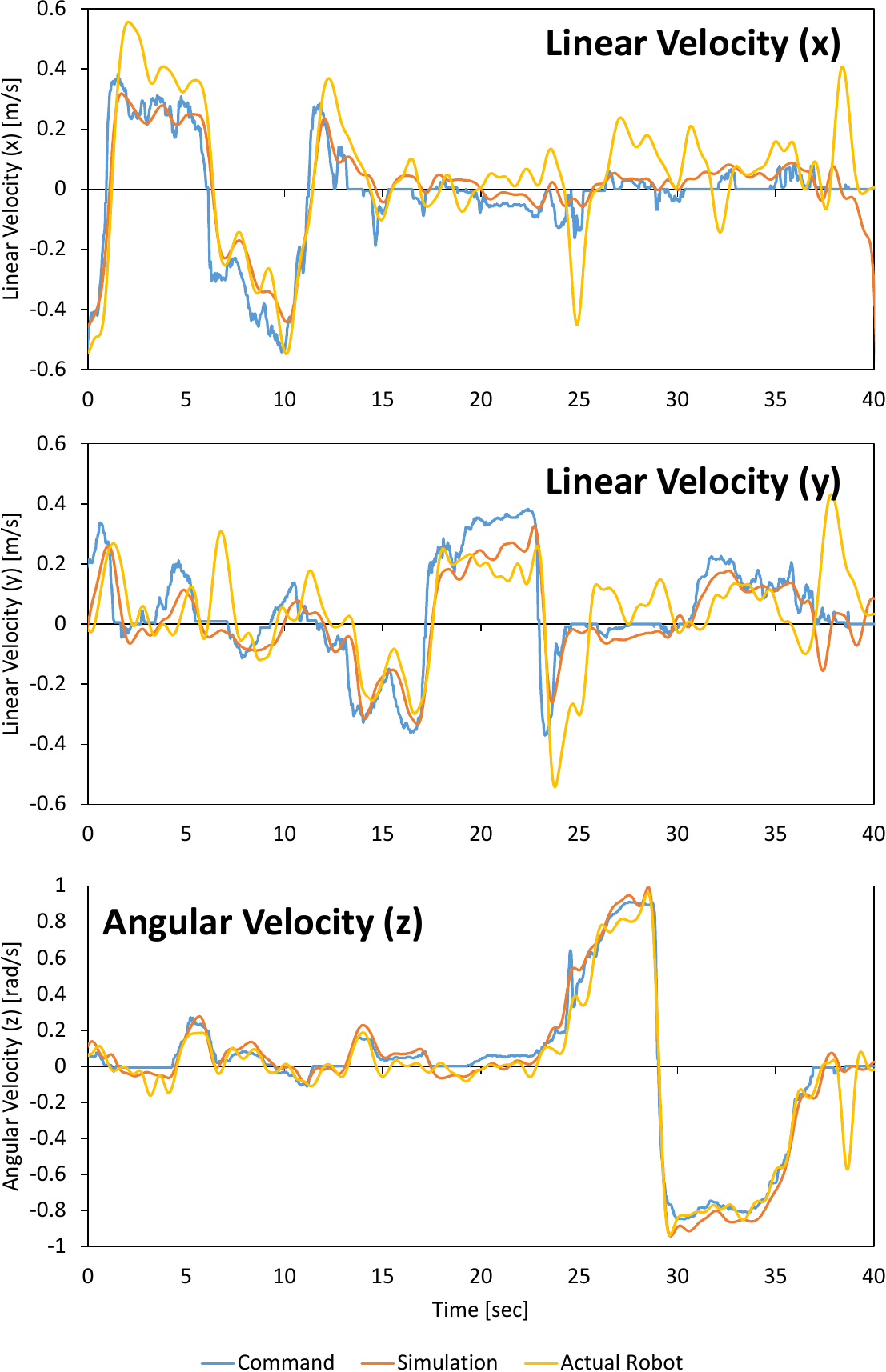}
  % \vspace{-1.0ex}
  \caption{Comparison of the tracking performance to the target commands when applying the trained policy to both MuJoCo simulation and the actual robot.}
  \label{figure:sim-act}
  % \vspace{-3.0ex}
\end{figure}

\begin{figure*}[t]
  \centering
  \includegraphics[width=1.95\columnwidth]{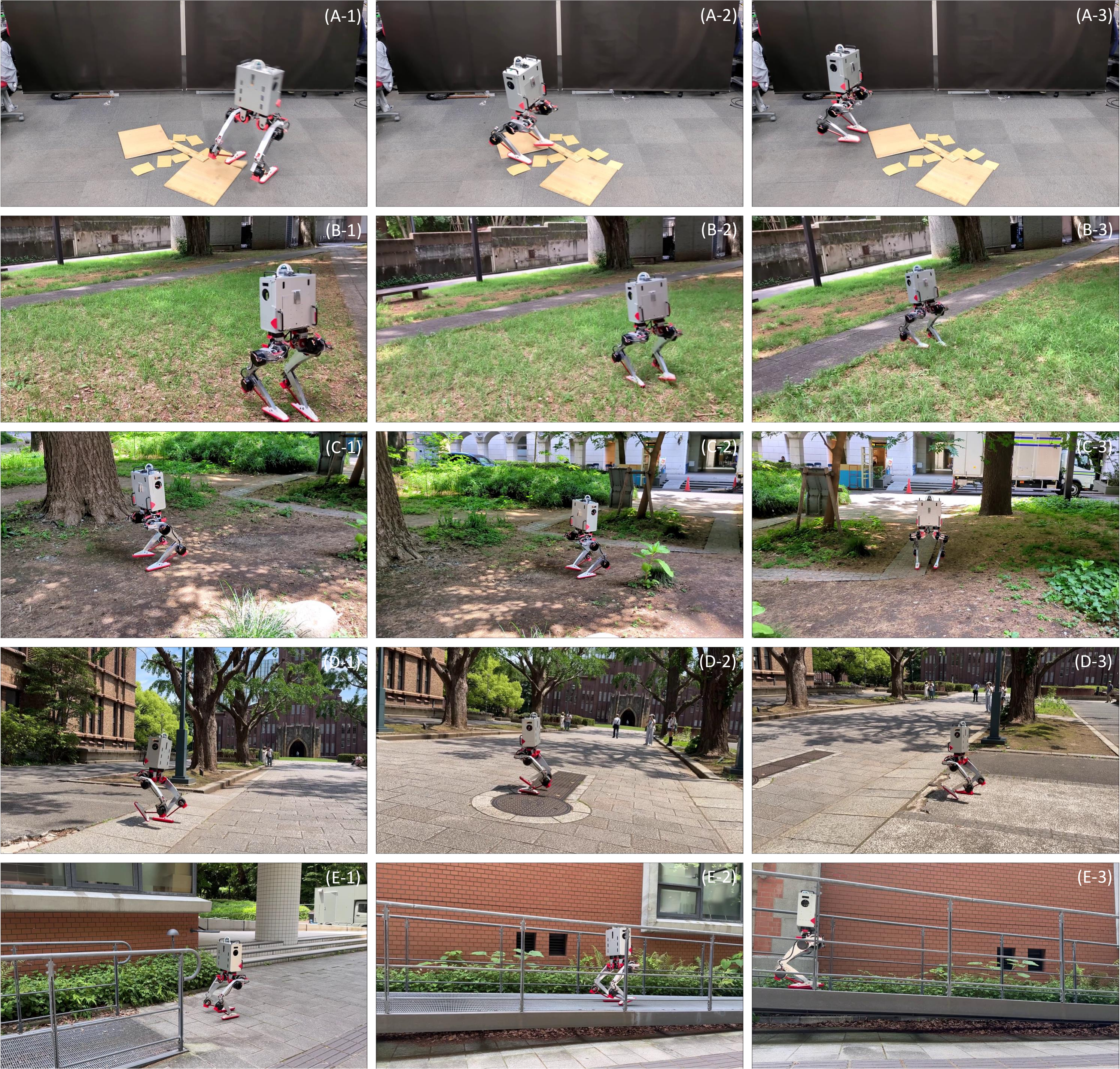}
  % \vspace{-1.0ex}
  \caption{Walking experiments in various environments: (A) uneven indoor terrain, (B) grassy field, (C) dirt surface, (D) concrete tiles, and (E) a gentle slope.}
  \label{figure:act-exp}
  % \vspace{-3.0ex}
\end{figure*}

\section{Experiments} \label{sec:experiment}

\switchlanguage%
{%
  First, we evaluated the differences observed when deploying a policy trained in IsaacGym to both MuJoCo simulation and the physical MEVITA robot.
  \figref{figure:sim-act} shows the tracking performance when the robot was commanded to follow linear and angular velocity inputs in forward, lateral, and rotational directions.
  The results indicate that the robot generally follows the commanded velocities and angular velocities in both simulation and the real world.
  In particular, the tracking of rotational commands shows minimal discrepancy between simulation and hardware.
  On the other hand, for translational velocity commands, the real robot exhibits slightly larger errors compared to the simulation.

  Next, \figref{figure:act-exp} shows walking experiments conducted in various environments.
  These include (A) uneven indoor terrain, (B) grass field, (C) dirt surface, (D) concrete tiles, and (E) a gentle slope.
  In all of these environments, MEVITA demonstrated stable walking behavior.

  On the other hand, it is difficult to say that the motion is completely stable.
  In particular, the robot is currently more unstable when standing still than when walking, indicating a need for policy adjustments.
  Additionally, excessive speed can lead to falls, and during lateral movements, the feet may get caught on the ground.
  Going forward, it will be necessary to carefully consider both Sim-to-Real and Real-to-Sim transfer to achieve more stable walking.
}%
{%
  まず, IsaacGymにより学習されたポリシーを, MuJoCoにおけるシミュレーションとロボット実機にデプロイした際の違いについて実験した.
  ロボットに前後左右回転の指令速度を与えたときの指令値への追従度合いを\figref{figure:sim-act}に示す.
  シミュレーションと実機ともに与えた指令速度・角速度に概ね追従していることがわかる.
  特に回転方向はシミュレーションと実機の差が小さい.
  これに対して, 並進方向の指令速度は, シミュレーションに比べて実機にはやや誤差が乗っていることがわかる.

  次に, 様々な環境における歩行実験の様子を\figref{figure:act-exp}に示す.
  (A)は室内の不整地, (B)は芝生, (C)は土の地面, (D)はコンクリートタイル, (E)は緩やかな傾斜で実験を行った様子である.
  いずれの環境においても, MEVITAは安定して歩行することができている.

  一方で, 動作が完全に安定しているとは言い難い.
  特に現状歩くよりも静止している状態のほうが不安定であり, ポリシーの調整が必要である.
  また, 速い速度を出しすぎると転びやすくなったり, 横移動を行うときに足が地面に引っかかったりすることもある.
  今後Sim-to-Real, Real-to-Simについて深く考慮し, より安定した歩行を実現する必要がある.
}%

\section{CONCLUSION} \label{sec:conclusion}
\switchlanguage%
{%
  In this study, we developed MEVITA, an open-source bipedal robot composed of metal components, all of which can be sourced from e-commerce platforms.
  One of the main challenges in metal-based open-source bipedal robots is the increased number of components due to the difficulty of fabricating freeform geometries.
  We addressed this by considering the minimal structural configuration and employing sheet metal welding to integrate multiple parts into single components.
  As a result, MEVITA achieves a significant reduction in the number of components compared to existing open-source bipedal robots, enabling an easily assemblable design that anyone can build with readily available parts.
  Despite its simple circuit architecture, MEVITA is capable of performing stable walking behaviors in various environments through reinforcement learning in simulation and Sim-to-Real transfer.
  By fully making all components open-source, we hope MEVITA will lower the barrier to bipedal robot development and contribute to the further advancement of research in physical intelligence.
}%
{%
  本研究では, 金属により構成され, ほぼ全ての部品をe-commerceサイトから構築できるオープンソース二足歩行ロボットMEVITAを開発した.
  金属製のオープンソース二足歩行ロボットは自由形状の難しさから部品点数が多くなってしまう問題を, 最小構成を考慮し, 板金溶接により複数の部品を一つにまとめることで解決した.
  これにより, 既存のオープンソース二足歩行ロボットよりも部品点数を大幅に削減し, 誰でも容易に部品を準備し組み立てられるような設計を実現した.
  本研究で開発したMEVITAはシンプルな回路構成ながら, シミュレーション上の強化学習とSim-to-Realにより, 様々な環境における歩行動作を実現可能であった.
  全ての部品を公開し, 誰でも簡単に二足歩行ロボットが構築できることで, physical intelligenceに関する様々な研究がより広がっていくことを期待している.
}%

{
  %\footnotesize
  %\small
  %\bibliographystyle{junsrt}
  \bibliographystyle{IEEEtran}
  \bibliography{main}
}

\end{document}